%% file: main_arxiv.tex
\documentclass[twoside,11pt]{article}

\usepackage{blindtext}

\usepackage[preprint]{jmlr2e}
\usepackage{graphicx}
\usepackage{caption}
\usepackage{subcaption}
\usepackage{placeins}
\usepackage{todonotes}
\usepackage{amsmath}
\usepackage{tikz}
\usepackage{float}
\usepackage{booktabs}       
\usetikzlibrary{patterns, shapes.geometric, positioning, arrows, shadows, matrix, fit, backgrounds}

\DeclareMathOperator*{\argmax}{arg\,max} 
\DeclareMathOperator*{\PCA}{PCA} 

\usepackage{lastpage}

\ShortHeadings{Causally Learning an Optimal Rework Policy}{O. Schacht et al.}
\firstpageno{1}

\begin{document}

\title{Causally Learning an Optimal Rework Policy}

\author{\name Oliver Schacht\footnotemark[1] \email oliver.schacht@uni-hamburg.de \\
    \name Sven Klaassen\footnotemark[1] \footnotemark[3]  \\
    \name Philipp Schwarz\footnotemark[1] \footnotemark[2]\\
	\name Martin Spindler\footnotemark[1] \footnotemark[3] \\
	\name Daniel Gr\"unbaum\footnotemark[2] \\
	\name Sebastiam Imhof\footnotemark[2] \\   
	\AND
       \addr \footnotemark[1] Faculty of Business Administration,
       University of Hamburg,
       Moorweidenstraße 18, 20148 Hamburg, Germany \\
       \addr \footnotemark[2] ams Osram,
       Leibnizstraße 4,
       93055 Regensburg, Germany \\
       \addr \footnotemark[3] Economic AI,
       Nürnberger Straße 262 A, 93059 Regensburg, Germany \\
       }

\maketitle

\begin{abstract}%
In manufacturing, rework refers to an optional step of a production process which aims to eliminate errors or remedy products that do not meet the desired quality standards. Reworking a production lot involves repeating a previous production stage with adjustments to ensure that the final product meets the required specifications. While offering the chance to improve the yield and thus increase the revenue of a production lot, a rework step also incurs additional costs. Additionally, the rework of parts that already meet the target specifications may damage them and decrease the yield. In this paper, we apply double/debiased machine learning (DML) to estimate the conditional treatment effect of a rework step during the color conversion process in opto-electronic semiconductor manufacturing on the final product yield. We utilize the implementation \texttt{DoubleML} to develop policies for the rework of components and estimate their value empirically. From our causal machine learning analysis we derive implications for the coating of monochromatic LEDs with conversion layers.
\end{abstract}


\input{sections/introduction.tex}

\begin{figure} 
\centering
  \includegraphics[width=0.8\textwidth]{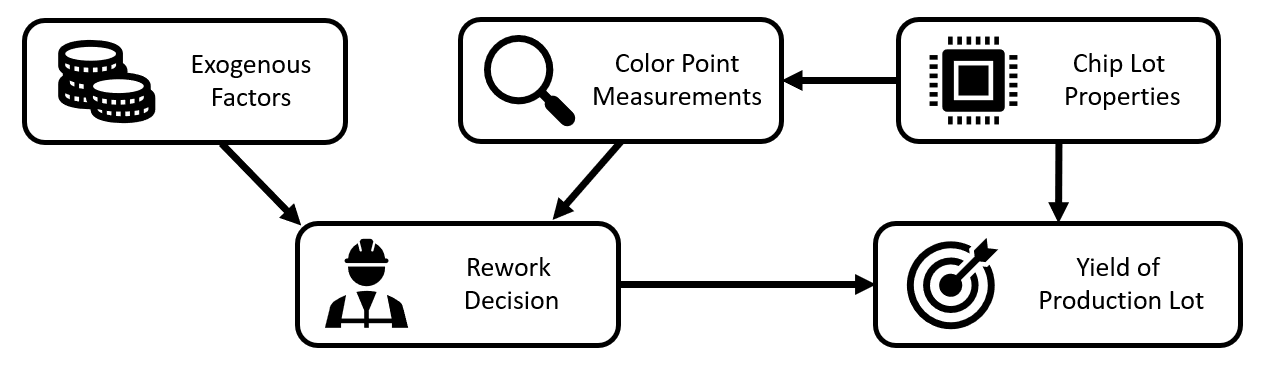}
  \caption{Causal Graph of the Rework Treatment Problem}
  \label{fig:graph}
\end{figure}

\input{sections/setting.tex}

\input{sections/methodology.tex}

\input{sections/results.tex}

\input{sections/conclusion.tex}

\begin{acks}
This work was funded by the Bavarian Joint Research Program (BayVFP) – Digitization (Funding reference: DIK0294/01). The research partners ams OSRAM and Economic AI kindly thank the the VDI/VDE-IT Munich for the organization and the Free State of Bavaria for the financial support. 
\end{acks}

\appendix

\input{sections/appendix/robustness_L3.tex}

\input{sections/appendix/results_L4.tex}

\bibliography{mybib}
\end{document}

%% file: sections/introduction.tex
\section{Introduction}\label{Sec:intro}
In the face of an ever-increasing pressure to produce high quality products at low costs, manufacturing companies are striving to transform products that do not meet the quality targets into sellable units by reworking them (\cite{liu2009}). For ecological reasons, too, reworking is preferable to discarding products that are not fulfilling preset target specifications.

The emergence of polychromatic light-emitting diodes (LEDs) extended the range of applications 
to areas in which traditional white light sources seemed predominant
(e.g. general home lighting, streetlamps, automotive, LCD displays), gradually superseding them.
Since then, multiple approaches for the production of white-emitting LEDs have been developed (for an in-depth introduction, see \cite{Cho2017}), with the most successful one being the coating of a monochromatic blue LED with a phosphor conversion layer.
The coating layer partially shifts the emitted wavelength spectrum of the light source,
resulting in a different perceived color (Figure \ref{intro:fig:ciexy}).
During manufacturing this so-called conversion step is crucial in order to produce LEDs with the color properties that meet the customers requirements.

In order to ensure a high yield, that is a high share of LEDs in a production lot that match the target color specifications, the rework workflow shown in Figure \ref{intro:fig:reworkflow} is used.
Starting at a mean color point \(C_0\) multiple conversion layers are applied to the chips of
a production lot, resulting in the converted mean color point \(C_1\).
Using the measurement \(C_1\) a decision is made, whether the chips are coated with an additional conversion layer.
After this optional rework step, production continues until the final products are assembled and tested.
Finally, the product specification is checked for each LED in the production lot using the test results 
(see mean color point \(C_2\) in Figure \ref{intro:fig:ciexy}).
Note that there are steps in between rework and the final testing, that have an influence on the final color point.
That is a certain color shift needs to be taken into account when applying the conversion material.

This process setup poses the question for which color points \(C_1\)
the additional rework step should be applied 
in order to improve the yield of a production lot.
The currently employed decision rule is based on the experience of the quality managers.

In the literature, rework policies are often determinated by methods of planning and control. For an overview of methodologies, see \citet{flapper2002}. The recent extension of machine learning (ML) from purely predictive tasks to causal questions enables the use of ML for learning decision policies. In this work, we are going to utilize double machine learning to derive rework policies based on real production data from the phosphorous conversion process of blue LEDs.

\begin{figure}
  \centering
  \begin{subfigure}[t]{.5\textwidth}
  \centering
    \scalebox{0.35}{\input{fig/CIExy.tex}}
  \caption{CIE 1931 color space chromaticity diagram}
    \label{intro:fig:ciexy}
  \end{subfigure}%
  \begin{subfigure}[t]{.5\textwidth}
    \centering
    \scalebox{0.85}{\input{fig/rework_flow.tex}}
    \caption{Simplified production workflow}
    \label{intro:fig:reworkflow}
  \end{subfigure}
  \caption{Conversion process including mean color point measurements before (\(C_0\)) and after conversion (\(C_1\)), as well as for the final LEDs (\(C_2\)) of a production lot.}
\end{figure}

\FloatBarrier

%% file: fig/CIExy.tex

\def\spectralLocus{ 
  (0.1738,0.0049)(0.1736,0.0049)(0.1733,0.0048)(0.1730,0.0048)(0.1726,0.0048)
  (0.1721,0.0048)(0.1714,0.0051)(0.1703,0.0058)(0.1689,0.0069)(0.1669,0.0086)
  (0.1644,0.0109)(0.1611,0.0138)(0.1566,0.0177)(0.1510,0.0227)(0.1440,0.0297)
  (0.1355,0.0399)(0.1241,0.0578)(0.1096,0.0868)(0.0913,0.1327)(0.0687,0.2007)
  (0.0454,0.2950)(0.0235,0.4127)(0.0082,0.5384)(0.0039,0.6548)(0.0139,0.7502)
  (0.0389,0.8120)(0.0743,0.8338)(0.1142,0.8262)(0.1547,0.8059)(0.1929,0.7816)
  (0.2296,0.7543)(0.2658,0.7243)(0.3016,0.6923)(0.3373,0.6589)(0.3731,0.6245)
  (0.4087,0.5896)(0.4441,0.5547)(0.4788,0.5202)(0.5125,0.4866)(0.5448,0.4544)
  (0.5752,0.4242)(0.6029,0.3965)(0.6270,0.3725)(0.6482,0.3514)(0.6658,0.3340)
  (0.6801,0.3197)(0.6915,0.3083)(0.7006,0.2993)(0.7079,0.2920)(0.7140,0.2859)
  (0.7190,0.2809)(0.7230,0.2770)(0.7260,0.2740)(0.7283,0.2717)(0.7300,0.2700)
  (0.7311,0.2689)(0.7320,0.2680)(0.7327,0.2673)(0.7334,0.2666)(0.7340,0.2660)
  (0.7344,0.2656)(0.7346,0.2654)(0.7347,0.2653)}

\def\planckianLocus{ 
  (0.6499,0.3474)(0.6361,0.3594)(0.6226,0.3703)(0.6095,0.3801)(0.5966,0.3887)
  (0.5841,0.3962)(0.5720,0.4025)(0.5601,0.4076)(0.5486,0.4118)(0.5375,0.4150)
  (0.5267,0.4173)(0.5162,0.4188)(0.5062,0.4196)(0.4965,0.4198)(0.4872,0.4194)
  (0.4782,0.4186)(0.4696,0.4173)(0.4614,0.4158)(0.4535,0.4139)(0.4460,0.4118)
  (0.4388,0.4095)(0.4320,0.4070)(0.4254,0.4044)(0.4192,0.4018)(0.4132,0.3990)
  (0.4075,0.3962)(0.4021,0.3934)(0.3969,0.3905)(0.3919,0.3877)(0.3872,0.3849)
  (0.3827,0.3820)(0.3784,0.3793)(0.3743,0.3765)(0.3704,0.3738)(0.3666,0.3711)
  (0.3631,0.3685)(0.3596,0.3659)(0.3563,0.3634)(0.3532,0.3609)(0.3502,0.3585)
  (0.3473,0.3561)(0.3446,0.3538)(0.3419,0.3516)(0.3394,0.3494)(0.3369,0.3472)
  (0.3346,0.3451)(0.3323,0.3431)(0.3302,0.3411)(0.3281,0.3392)(0.3261,0.3373)
  (0.3242,0.3355)(0.3223,0.3337)(0.3205,0.3319)(0.3188,0.3302)(0.3171,0.3286)
  (0.3155,0.3270)(0.3140,0.3254)(0.3125,0.3238)(0.3110,0.3224)(0.3097,0.3209)
  (0.3083,0.3195)(0.3070,0.3181)(0.3058,0.3168)(0.3045,0.3154)(0.3034,0.3142)
  (0.3022,0.3129)(0.3011,0.3117)(0.3000,0.3105)(0.2990,0.3094)(0.2980,0.3082)
  (0.2970,0.3071)(0.2961,0.3061)(0.2952,0.3050)(0.2943,0.3040)(0.2934,0.3030)
  (0.2926,0.3020)(0.2917,0.3011)(0.2910,0.3001)(0.2902,0.2992)(0.2894,0.2983)
  (0.2887,0.2975)(0.2880,0.2966)(0.2873,0.2958)(0.2866,0.2950)(0.2860,0.2942)
  (0.2853,0.2934)(0.2847,0.2927)(0.2841,0.2919)(0.2835,0.2912)(0.2829,0.2905)
  (0.2824,0.2898)(0.2818,0.2891)(0.2813,0.2884)(0.2807,0.2878)(0.2802,0.2871)
  (0.2797,0.2865)(0.2792,0.2859)(0.2788,0.2853)(0.2783,0.2847)(0.2778,0.2841)
  (0.2774,0.2836)(0.2770,0.2830)(0.2765,0.2825)(0.2761,0.2819)(0.2757,0.2814)
  (0.2753,0.2809)(0.2749,0.2804)(0.2745,0.2799)(0.2742,0.2794)(0.2738,0.2789)
  (0.2734,0.2785)(0.2731,0.2780)(0.2727,0.2776)(0.2724,0.2771)(0.2721,0.2767)
  (0.2717,0.2763)(0.2714,0.2758)(0.2711,0.2754)(0.2708,0.2750)(0.2705,0.2746)
  (0.2702,0.2742)(0.2699,0.2738)(0.2696,0.2735)(0.2694,0.2731)(0.2691,0.2727)
  (0.2688,0.2724)(0.2686,0.2720)(0.2683,0.2717)(0.2680,0.2713)(0.2678,0.2710)
  (0.2675,0.2707)(0.2673,0.2703)(0.2671,0.2700)(0.2668,0.2697)(0.2666,0.2694)
  (0.2664,0.2691)(0.2662,0.2688)(0.2659,0.2685)(0.2657,0.2682)(0.2655,0.2679)
  (0.2653,0.2676)(0.2651,0.2673)(0.2649,0.2671)(0.2647,0.2668)(0.2645,0.2665)
  (0.2643,0.2663)(0.2641,0.2660)(0.2639,0.2657)(0.2638,0.2655)(0.2636,0.2652)
  (0.2634,0.2650)(0.2632,0.2648)(0.2631,0.2645)(0.2629,0.2643)(0.2627,0.2641)
  (0.2626,0.2638)(0.2624,0.2636)(0.2622,0.2634)(0.2621,0.2632)(0.2619,0.2629)
  (0.2618,0.2627)(0.2616,0.2625)(0.2615,0.2623)(0.2613,0.2621)(0.2612,0.2619)
  (0.2610,0.2617)(0.2609,0.2615)(0.2608,0.2613)(0.2606,0.2611)(0.2605,0.2609)
  (0.2604,0.2607)(0.2602,0.2606)(0.2601,0.2604)(0.2600,0.2602)(0.2598,0.2600)
  (0.2597,0.2598)(0.2596,0.2597)(0.2595,0.2595)(0.2593,0.2593)(0.2592,0.2592)
  (0.2591,0.2590)(0.2590,0.2588)(0.2589,0.2587)(0.2588,0.2585)(0.2587,0.2584)
  (0.2586,0.2582)(0.2584,0.2580)(0.2583,0.2579)(0.2582,0.2577)(0.2581,0.2576)
  (0.2580,0.2574)(0.2579,0.2573)(0.2578,0.2572)(0.2577,0.2570)(0.2576,0.2569)
  (0.2575,0.2567)(0.2574,0.2566)(0.2573,0.2565)(0.2572,0.2563)(0.2571,0.2562)
  (0.2571,0.2561)(0.2570,0.2559)(0.2569,0.2558)(0.2568,0.2557)(0.2567,0.2555)
  (0.2566,0.2554)(0.2565,0.2553)(0.2564,0.2552)(0.2564,0.2550)(0.2563,0.2549)
  (0.2562,0.2548)(0.2561,0.2547)(0.2560,0.2546)(0.2559,0.2545)(0.2559,0.2543)
  (0.2558,0.2542)(0.2557,0.2541)(0.2556,0.2540)(0.2556,0.2539)(0.2555,0.2538)
  (0.2554,0.2537)(0.2553,0.2536)(0.2553,0.2535)(0.2552,0.2534)(0.2551,0.2533)
  (0.2551,0.2532)(0.2550,0.2531)(0.2549,0.2530)(0.2548,0.2529)(0.2548,0.2528)
  (0.2547,0.2527)(0.2546,0.2526)(0.2546,0.2525)(0.2545,0.2524)(0.2544,0.2523)
  (0.2544,0.2522)(0.2543,0.2521)(0.2543,0.2520)(0.2542,0.2519)(0.2541,0.2518)
  (0.2541,0.2517)(0.2540,0.2516)(0.2540,0.2516)(0.2539,0.2515)(0.2538,0.2514)
  (0.2538,0.2513)(0.2537,0.2512)(0.2537,0.2511)(0.2536,0.2511)(0.2535,0.2510)
  (0.2535,0.2509)(0.2534,0.2508)(0.2534,0.2507)(0.2533,0.2507)(0.2533,0.2506)
  (0.2532,0.2505)(0.2532,0.2504)(0.2531,0.2503)(0.2531,0.2503)(0.2530,0.2502)
  (0.2530,0.2501)(0.2529,0.2500)(0.2529,0.2500)(0.2528,0.2499)(0.2528,0.2498)
  (0.2527,0.2497)(0.2527,0.2497)(0.2526,0.2496)(0.2526,0.2495)(0.2525,0.2495)
  (0.2525,0.2494)(0.2524,0.2493)(0.2524,0.2493)(0.2523,0.2492)(0.2523,0.2491)
  (0.2523,0.2491)(0.2522,0.2490)(0.2522,0.2489)(0.2521,0.2489)(0.2521,0.2488)
  (0.2520,0.2487)(0.2520,0.2487)(0.2519,0.2486)(0.2519,0.2485)(0.2519,0.2485)
  (0.2518,0.2484)(0.2518,0.2484)(0.2517,0.2483)(0.2517,0.2482)(0.2517,0.2482)
  (0.2516,0.2481)(0.2516,0.2481)(0.2515,0.2480)(0.2515,0.2480)(0.2515,0.2479)
  (0.2514,0.2478)(0.2514,0.2478)(0.2513,0.2477)(0.2513,0.2477)(0.2513,0.2476)
  (0.2512,0.2476)(0.2512,0.2475)(0.2512,0.2474)(0.2511,0.2474)(0.2511,0.2473)
  (0.2511,0.2473)(0.2510,0.2472)(0.2510,0.2472)(0.2509,0.2471)(0.2509,0.2471)
  (0.2509,0.2470)(0.2508,0.2470)(0.2508,0.2469)(0.2508,0.2469)(0.2507,0.2468)
  (0.2507,0.2468)(0.2507,0.2467)(0.2506,0.2467)(0.2506,0.2466)(0.2506,0.2466)
  (0.2505,0.2465)(0.2505,0.2465)(0.2505,0.2464)(0.2505,0.2464)(0.2504,0.2463)
  (0.2504,0.2463)(0.2504,0.2463)(0.2503,0.2462)(0.2503,0.2462)(0.2503,0.2461)
  (0.2502,0.2461)(0.2502,0.2460)(0.2502,0.2460)(0.2502,0.2459)(0.2501,0.2459)
  (0.2501,0.2459)(0.2501,0.2458)(0.2500,0.2458)(0.2500,0.2457)(0.2500,0.2457)
  (0.2500,0.2456)(0.2499,0.2456)(0.2499,0.2456)(0.2499,0.2455)(0.2498,0.2455)
  (0.2498,0.2454)(0.2498,0.2454)(0.2498,0.2454)(0.2497,0.2453)(0.2497,0.2453)
  (0.2497,0.2452)(0.2497,0.2452)(0.2496,0.2452)(0.2496,0.2451)(0.2496,0.2451)
  (0.2496,0.2450)(0.2495,0.2450)(0.2495,0.2450)(0.2495,0.2449)(0.2495,0.2449)
  (0.2494,0.2449)(0.2494,0.2448)(0.2494,0.2448)(0.2494,0.2447)(0.2493,0.2447)
  (0.2493,0.2447)(0.2493,0.2446)(0.2493,0.2446)(0.2492,0.2446)(0.2492,0.2445)
  (0.2492,0.2445)(0.2492,0.2445)(0.2491,0.2444)(0.2491,0.2444)(0.2491,0.2444)
  (0.2491,0.2443)(0.2491,0.2443)(0.2490,0.2443)(0.2490,0.2442)(0.2490,0.2442)
  (0.2490,0.2442)(0.2489,0.2441)(0.2489,0.2441)(0.2489,0.2441)(0.2489,0.2440)
  (0.2489,0.2440)(0.2488,0.2440)(0.2488,0.2439)(0.2488,0.2439)(0.2488,0.2439)
  (0.2487,0.2438)}



\def\primariesLoci{
  (0.6400,0.3300)  
  (0.3000,0.6000)  
  (0.1500,0.0600)} 

\def\whitepointLocus{
  (0.3127,0.3290)}

\def\XYZtoRGB{
  { 3.2410}{-1.5374}{-0.4986}
  {-0.9692}{ 1.8760}{ 0.0416}
  { 0.0556}{-0.2040}{ 1.0570}}

\def\gammaCorrect{
  dup 0.0031308 le                    
  {12.92 mul}                         
  {1 2.4 div exp 1.055 mul -0.055 add}
  ifelse }


\def\scalarProduct#1#2#3{
  #3 mul     exch
  #2 mul add exch
  #1 mul add }

\def\applyMatrix#1#2#3#4#5#6#7#8#9{
  3 copy 3 copy
  \scalarProduct{#7}{#8}{#9} 7 1 roll
  \scalarProduct{#4}{#5}{#6} 5 1 roll
  \scalarProduct{#1}{#2}{#3} 3 1 roll }

\def\xyYtoXYZ{                        
  3 copy 3 1 roll                     
  add neg 1 add mul 2 index div       
  4 1 roll                            
  dup                                 
  5 1 roll                            
  3 2 roll                            
  mul exch div                        
  3 1 roll }                          

\def\gammaCorrectVector{
  \gammaCorrect 3 1 roll
  \gammaCorrect 3 1 roll
  \gammaCorrect 3 1 roll}

\begin{tikzpicture}

\pgfdeclarefunctionalshading{colorspace}
  {\pgfpointorigin}{\pgfpoint{1000bp}{1000bp}}{}{
    1000 div exch 1000 div exch       
    1.0                               
    \xyYtoXYZ                         
    \expandafter\applyMatrix\XYZtoRGB 
    \gammaCorrectVector }             

\begin{scope} [shift={(-500bp,-500bp)}, scale=100bp/2cm]
  \fill [white] (-1,-1) rectangle (9,10);
  \draw [dashed, gray] grid (8,9);
  \begin{scope} [scale=10]
    \path [mark=*, mark repeat=2, white] 
      plot [mark size=0.10, mark phase=1] coordinates {\spectralLocus}
      plot [mark size=0.05, mark phase=2] coordinates {\spectralLocus};
    \clip [smooth] plot coordinates {\spectralLocus} -- cycle;
     \node[anchor=south west,%
        inner sep=0,%
        scale=2,%
        outer sep=0pt] (image) at (0, 0) {\includegraphics{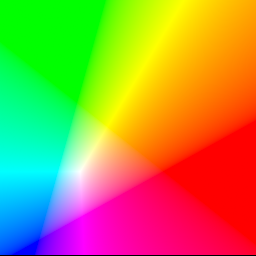}};
    \fill [black, even odd rule, opacity=0.1]
      rectangle +(1,1) plot coordinates {\primariesLoci} -- cycle;
  \end{scope}

  \begin{scope} [
    scale=10,
    every node/.style = {circle,draw,fill=white,minimum width=0.2cm}
  ]
    \draw[ultra thick, black, loosely dotted] plot [smooth, tension=1] 
      coordinates { (0.16, 0.08) (0.38, 0.39) (0.48, 0.41)};

    \node[label=right:{\Huge\(C_0\)},] (START) at (0.16, 0.08)  {};
    \node[label=right:{\Huge\(C_1\)},] (CONV) at (0.28, 0.264)  {};
    \node[label=right:{\Huge\(C_2\)},] (END) at (0.35, 0.315)  {};

  \end{scope}

  \begin{scope} [
    scale=10,
    every node/.style = {circle}
  ]
    \node (START) at (0.4, -0.04)  {\Huge\(C_x\)};
    \node (START) at (-0.04, 0.45)  {\Huge\(C_y\)};
  \end{scope}
\end{scope}
\end{tikzpicture}

%% file: fig/rework_flow.tex
\begin{tikzpicture}
  \begin{scope}[
    thick,
    every node/.style={minimum width=2.6cm, minimum height=0.8cm},
    decision/.style={diamond,draw,top color=white,bottom color=yellow!20,,general shadow={fill=gray!60,shadow xshift=1pt,shadow yshift=-1pt}},
    block/.style={rectangle,draw,top color=white,bottom color=blue!10,general shadow={fill=gray!60,shadow xshift=1pt,shadow yshift=-1pt}},
    hidden/.style={rectangle,fill=white,text opacity=1,fill opacity=0},
  ]
    \node[block] (CONV) { apply converter };
    \node[block,below=0.5cm of CONV] (ICC) { measure (\(C_1\)) };
    \node[decision,aspect=2,below=0.5cm of ICC] (DECISION) {colour reached?};
    \node[block,right=of DECISION] (REW) { rework };
    \node[hidden,below=0.5cm of DECISION] (OTHER) { \(\ldots\) };
    \node[block,below=0.5cm of OTHER] (FT) { final test (\(C_2\)) };

    \path [->] (CONV) edge (ICC);
    \path [->] (ICC) edge (DECISION);
    \path [->] (DECISION) edge node[midway,above] {no} (REW);
    \path [->] (DECISION) edge node[midway,left] {yes} (OTHER);
    \path [draw, ->] (REW) |- (OTHER);
    \path [->] (OTHER) edge (FT);
  \end{scope}
\end{tikzpicture}

%% file: sections/setting.tex
\section{Setting}
\subsection{Data} \label{Sec:Data}
The collected data consists of observations $W_i=\{Y_i,A_i,X_i\}$, where $Y_i$ denotes the outcome, which is a measure of the final yield of each production lot.
Further, $A_i \in \{0,1\}$ represents the treatment/action variable, corresponding to the decision whether to rework a production lot and $X_i$ are characteristics of the production lot before assignment of a rework.
Finally, we have data of two different product types (\textit{P1} and \textit{P2}) with $n_{P1}=32,669$ and $n_{P2}=17,802$ observations.
In the observed data, there is no clear policy in place.
The quality managers assign the rework treatment by visually inspecting the plot of the covariates 
$X_i = \{(C_{1,x}^i), (C_{1,y}^i)\}$, consisting of the CIE coordinates of the mean color point \(C_1^i\) after the conversion.
Thus, under the current policy, there is an overlap of production lots with similar characteristics after conversion but with different treatments assigned. 

As indicated by the color points \(C_0\), \(C_1,\) and \(C_2\) in Figure \ref{intro:fig:ciexy} 
the application of the conversion material shifts the resulting color point along a curve
in the CIE color space from an unconverted monochromatic emission (\(C_0\)) to a fully converted one (\(C_2\)).
Thereby, the amount of applied conversion material controls the position on this \textit{conversion curve}.
Minor deviations orthogonal to the conversion curve can be explained through process instabilities.
Due to the scale of the $C_1$ measurements, the conversion curve can be well approximated by a linear function.
Thus, application of a principal component analysis (PCA) leads to transformed color coordinates 
\(\PCA(C_1) = (C_m, \, C_s)\) where the first principal component \(C_m\) captures the
position on the conversion curve representing the main decision criteria,
and the second principal component \(C_s\) is mainly determined by process fluctuations.
In the following we refer to \(C_m\) as the \textit{main color point measure} and to \(C_s\) as the \textit{secondary color point measure}.

Figure \ref{fig:graph} represents the assumptions on the underlying causal structure in form of a directed acyclic graph (DAG, see \cite{pearl1995}). The yield of a production lot depends to a large extent on how close the chips are to the target color location in the final quality control. As the rework decision is based on the mean color point measurements after phosphorous conversion, the luminous properties of the opto-electronic chips influence not only the yield but also the decision to rework a specific product lot. Consequently, the luminous properties act as confounders with respect to the rework effect. Although there are other influencing factors on the treatment decision (i.e. the individual employee) or the outcome (i.e. variations in downstream process steps), these do not affect both simultaneously. 

\subsection{Effect Identification}
As argued in Section \ref{Sec:Data} the rework decision is done based on the characteristics $X$. Consequently, conditioning on $X$ will close all backdoor paths between rework decision and yield. Put differently, we base our identification on conditional exogeneity, such that
$$Y(a)\perp A |X,$$
where $Y(a)$ denotes the potential outcome for $a\in \{0,1\}$ (see \cite{rubin2005}).
Under the assumptions above, the underlying causal structure can be represented by an interactive regression model (IRM):
\begin{align*}
    Y &= g_0(A,X) + U,\quad \mathbb{E}[U\mid X,A] =0,\\
    A &= m_0(X) + V,\quad \mathbb{E}[V\mid X] = 0,
\end{align*}
where the conditional expectations 
\begin{align}
    g_0(A,X) &= \mathbb{E}[Y\mid X, A],\label{eq:g_0}\\
    m_0(X) &= \mathbb{E}[A\mid X] = P(A=1\mid X)\label{eq:m_0}
\end{align}
are unknown and might be complex functions of $X$. In this structural equation model, the average treatment effect (ATE)
\begin{equation*}\label{ATE}
    \theta_{ATE} = \mathbb{E}[g_0(1,X)-g_0(0,X)]
\end{equation*}
as well as the average treatment effect of the treated (ATTE)
\begin{equation*}\label{ATTE}
    \theta_{ATTE} = \mathbb{E}[g_0(1,X)-g_0(0,X)\mid A=1]
\end{equation*}
are identified. The work of \citet{chernozhukov2018} enables the use of machine learning (ML) algorithms such as random forest or boosting to obtain precise estimates of treatment effects combined with confidence intervals to access estimation uncertainty.

In double machine learning, inference is based on a method of moments estimator
\begin{equation}
    E[\psi(W;\theta_0,\eta_0)] = 0,
\end{equation}
with $\psi(W;\theta_0,\eta_0)$ being a Neyman-orthogonal score that identifies the causal parameter $\theta_0$ given ML estimates of the parameters or functions $\eta_0$ (referred to as nuisance). The Neyman-orthogonality property ensures robustness of the estimator against small perturbations in the ML nuisance estimates. Further, the usage of $k$-fold crossfitting safeguards against overfitting and enables to control the complexity of the estimator, leading to appealing properties such as $\sqrt{n}$-consistency and approximate normality.

In case of an IRM, the score for an ATE estimator is given by the linear form
\begin{align}
	\psi(W_i;\theta,\eta) :=&\ \psi_a(W_i,\eta)\theta + \psi_b(W_i,\eta)\\
	=&\ - \theta + g(1,X_i) - g(0,X_i) \nonumber \\
	 &\ + \frac{A_i(Y_i - g(1,X_i))}{m(X_i)} - \frac{(1-A_i)(Y_i - g(0,X_i))}{1-m(X_i)},
\label{eq:psi}
\end{align}
which is also known as augmented inverse propensity weighting (AIPW) (\cite{robins1995}).

\begin{figure}[h]
    \centering
    \includegraphics[width=0.6\linewidth]{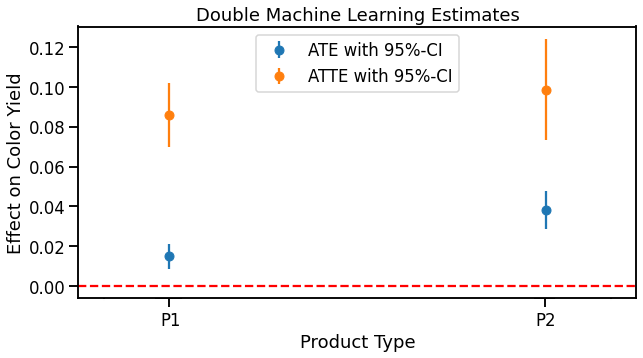}
    \caption{Average treatment effect of reworking a chip lot and average treatment effect on the treated estimated by double machine learning.}
    \label{fig:ATE}
\end{figure}

Applied to the data, the estimates (see Figure \ref{fig:ATE}) suggest that for both products, the treatment effect is much higher on the treated than on the whole population. This emphasizes that treating all individual lots is not desirable in this setting, as the effect varies over the covariates $X$. \\

\begin{figure}[h]
    \centering
    \includegraphics[width=\linewidth]{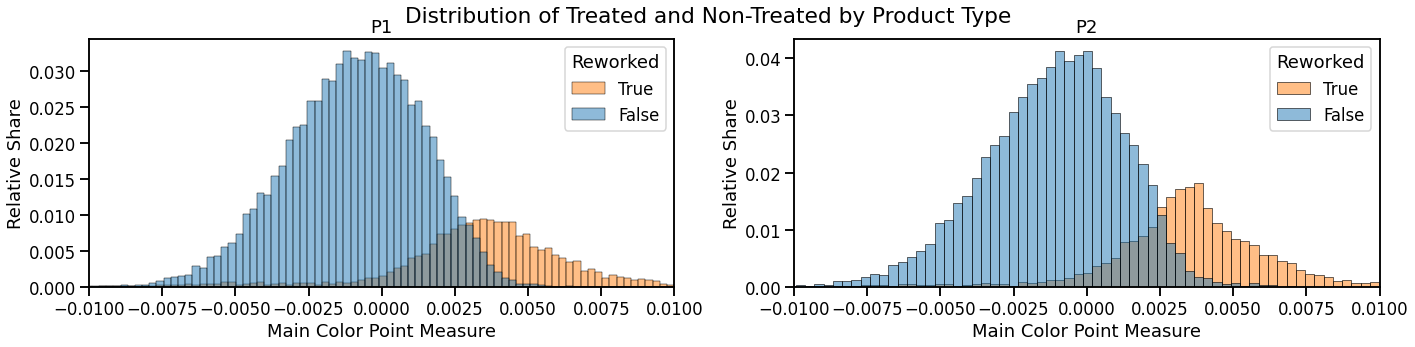}
    \caption{Distribution of the main color point measurement. Observations with a low value are already close to the target, thus they have a low treatment probability. Observations with a high value are therefore treated with a higher probability.}
    \label{fig:behavpol}
\end{figure}

Figure \ref{fig:behavpol} shows the distribution of the main color point measure in $X$ for the treated and non-treated subsamples. There is no clear cut-off due to the reasons mentioned above. Ideally, there would be a policy in place that assigns the treatment to each production lot if the corresponding effect is above a set threshold (incorporating costs etc.). Our goal is to estimate a policy $\pi\in\Pi$, that maps individuals features $X_i \in \mathcal{X}$ to a treatment decision: $\pi : \mathcal{X} \rightarrow \{0, 1\}$, which incorporates the heterogeneity of the effect of the rework.

During the rest of the main analysis, for simplicity, we focus on the product type P1 (results for P2 can be found in Appendix \ref{app:L4}).

%% file: sections/methodology.tex
\section{Methodology}
As motivated in the previous section, in this scenario the treatment effect is heterogeneous with respect to covariates $X$. We will employ different approaches from conditional treatment effect estimation and optimal policy learning to learn a rework policy in this setting.

\subsection{Conditional Average Treatment Effects} \label{Sec:CATE}
Given a set of the covariates $\tilde{X}$ (not necessarily included in $X$), the CATE is defined as
$$\theta_0(\tilde{x}):=\mathbb{E}[g_0(1,X) - g_0(0,X)|\tilde{X}=\tilde{x}].$$
\citet{semenova2021debiased} propose to approximate $\theta_0(\tilde{x})\approx b(\tilde{x})^T\beta$ via a linear form, where $b(\tilde{x})$ is a $d$-dimensional basis vector of $\tilde{x}$. 
The idea is based on projecting the part $\psi_b(W_i,\eta)$ of the Neyman-orthogonal scores onto the predefined basis vector $b(\tilde{x})$.
The authors provide extensive theory to construct pointwise and jointly valid confidence intervals via gaussian approximations.
As explained in Section \ref{Sec:Data}, being close to the color target might be well approximated via the first principal component of the production lot properties $X$. 

\begin{figure}[ht]
    \centering
    \includegraphics[width=\linewidth]{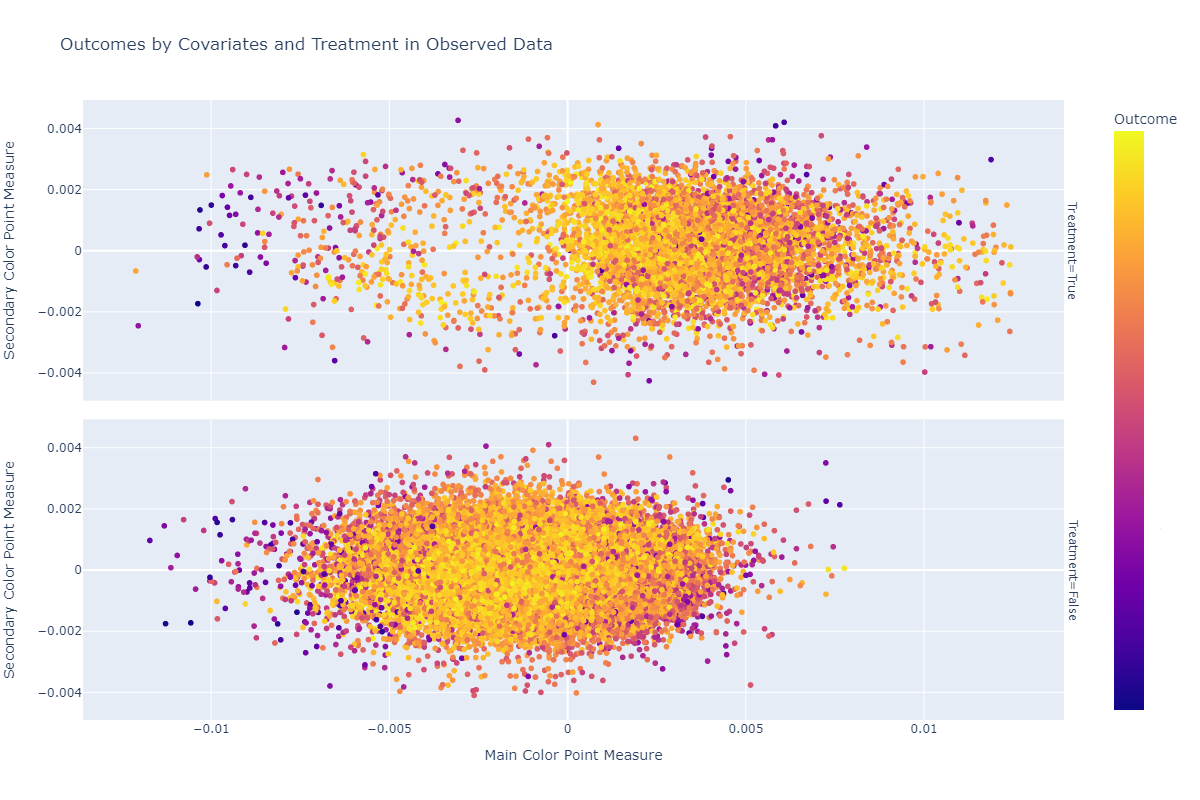}
    \caption{Principal components of the chip properties before treatment, treatment decision and observed outcome in the data for product type ``P1''. }
    \label{fig:pca_obs}
\end{figure}

\begin{figure}[ht]
    \centering
    \includegraphics[width=0.75\linewidth]{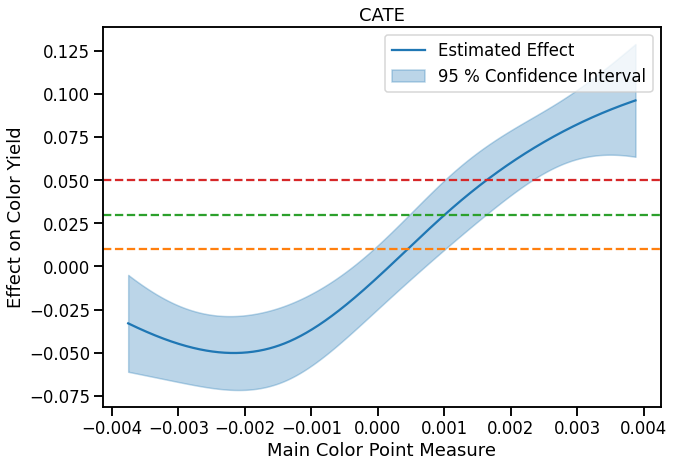}
    \caption{Estimation of the CATE along the first color point measurement. The horizontal dashed lines represent different policy thresholds $\gamma$.}
    \label{fig:cate_1d}
\end{figure}

To access the heterogeneity associated with the first principal component, we set $\tilde{X}$ to be the first principal component of the covariates $X$. Further, to allow for a flexible conditional treatment effect, we use a b-spline basis for $b(\tilde{x})$. The effect estimates combined with confidence intervals are displayed in Figure \ref{fig:cate_1d}.

As a positive CATE indicates improvements in yield, our estimated policy will be based on whether the effect on the yield surpasses some given threshold

$$\hat{\theta}(\tilde{x}) \ge \gamma,$$

where $\tilde{x}$ is the first principal component of $X$ and the threshold $\gamma>0$ can be used to incorporate costs and select rework groups at different effect levels.

\begin{figure}[ht]
    \centering
    \includegraphics[width=0.8\linewidth]{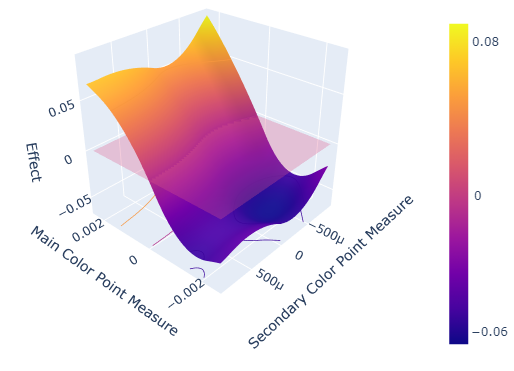}
    \caption{CATE with respect to both color point measurements.}
    \label{fig:cate_2d}
\end{figure}

Additionally considering the second principal component can contain further information about treatment effect heterogeneity and might improve the resulting policy. To be able to approximate the conditional effect flexibly based on the first two principal components, we rely on a tensor product of b-splines to construct our basis vector $b(\tilde{x})$. The resulting effect estimates are displayed in Figure \ref{fig:cate_2d}, where we omitted the confidence regions for clarity.

It is visible that the first principal component captures the main heterogeneity of the treatment effect. Nevertheless, we will compare the difference in the policies in Section \ref{Sec:results}.

In both cases above, instead of using the effect estimate, using the lower bound of the confidence interval, will result in more conservative policy but account for the estimation uncertainty (see Appendix \ref{sec:confint}).
Additionally, one might consider the percentage of the reworked panels, to decide which threshold might be preferable to incorporate capacity constraints.

\subsection{Policy Leaning}
An alternative approach is to directly estimate a policy based on the part $\psi_b(W_i,\eta)$ in Equation \ref{eq:psi} as introduced by \citet{athey2021}. They propose to estimate the treatment assignment rule as
\begin{equation}
	\hat{\pi}=\argmax_{\pi\in\Pi}\frac{1}{n}\sum_{i=1}^{n}(2\pi(X_i)-1)\psi_b(W_i,\hat{\eta}).
	\label{eq:policy}
\end{equation}
Given a specified policy class $\Pi$ and the corresponding regret
\begin{align*}
R(\pi):=\max_{\pi'\in\Pi}\left\{\mathbb{E}[Y_i(\pi'(X_i))]\right\} - \mathbb{E}[Y_i(\pi(X_i))],
\end{align*}
the authors are able to derive regret bounds of order $1/\sqrt{n}$. 

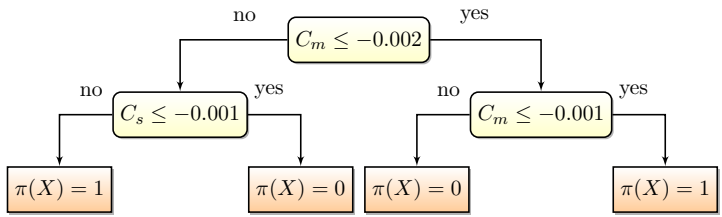
\begin{figure}[h!]
    \centering
    \scalebox{0.75}{\input{fig/example_tree.tex}}
    \caption{Exemplary policy tree of depth $2$.}
    \label{fig:tree}
\end{figure}

The learning problem in Equation \ref{eq:policy} can be reformulated as a weighted classification problem
\begin{equation}
	\hat{\pi}=\argmax_{\pi\in\Pi}\frac{1}{n}\sum_{i=1}^{n}\lambda_i H_i(2\pi(X_i)-1),
\end{equation}
with weights $\lambda_i=|\psi_b(W_i,\hat{\eta})|$ and target $H_i = \textrm{sign}(\psi_b(W_i,\hat{\eta}))$. Thus, we define the space of possible policies $\Pi$ to be the class of depth-$m$ decision trees (see Figure \ref{fig:tree} as an example) with $m \in \{1,2\}$. Appendix \ref{app:rob} includes results for $\Pi$ being the class of linear decision boundaries after application of the radial basis function (rbf) kernel (using support vector machines).
To evaluate policies at different thresholds, we reduce the score by the desired threshold $\psi_b(W_i,\hat{\eta})-\gamma$ before classification.

%% file: fig/example_tree.tex
\begin{tikzpicture}
  \begin{scope}[
    thick,
    every node/.style={minimum height=0.8cm},
    decision/.style={rectangle,draw,rounded corners,top color=white,bottom color=yellow!20,general shadow={fill=gray!60,shadow xshift=1pt,shadow yshift=-1pt}},
    block/.style={rectangle,draw,top color=white,bottom color=orange!40,general shadow={fill=gray!60,shadow xshift=1pt,shadow yshift=-1pt}},
    hidden/.style={rectangle,fill=white,text opacity=1,fill opacity=0},
    line/.style={draw,-latex'},
  ]
    \node[decision] (TOP) { \(C_m \leq -0.002\) };
    \node[decision, below left = 0.5 and 0.7 of TOP] (MID1) { \(C_s \leq -0.001\) };
    \node[decision, below right = 0.5 and 0.7 of TOP] (MID2) { \(C_m \leq -0.001\) };

    \node[block, below left = 0.5 and 0 of MID1] (END1) { \(\pi(X) = 1\) };
    \node[block, below right = 0.5 and 0 of MID1] (END2) { \(\pi(X) = 0\) };
    \node[block, below left = 0.5 and 0 of MID2] (END3) { \(\pi(X) = 0\) };
    \node[block, below right = 0.5 and 0 of MID2] (END4) { \(\pi(X) = 1\) };


    \path [line] (TOP) -| node[above, pos=0.2] {no} (MID1);
    \path [line] (TOP) -| node[above, pos=0.2] {yes} (MID2);

    \path [line] (MID1) -| node[above, pos=0.2] {no} (END1);
    \path [line] (MID1) -| node[above, pos=0.2] {yes} (END2);

    \path [line] (MID2) -| node[above, pos=0.2] {no} (END3);
    \path [line] (MID2) -| node[above, pos=0.2] {yes} (END4);
  \end{scope}
\end{tikzpicture}

%% file: sections/results.tex
\section{Results}\label{Sec:results}

The implementation of the policy estimation relies on the \texttt{DoubleML}-package (\cite{bach2022doubleml, bach2021doubleml}). The package allows for direct estimation of the IRM and the CATEs with a variety of machine learning algorithms.
The implemented \texttt{DoubleML} model is based on the original data without principal component transformation. As the quality of the policy estimation will improve with high-quality estimates of the unknown nuisance elements $\eta=(g_0, m_0)$ (see \eqref{eq:g_0} and \eqref{eq:m_0}), we rely on tuned machine learning algorithms. Further, we employ $5$-fold crossfitting and trim the propensity score estimate $\hat{m}$ at $0.025$ and $0.975$, respectively. As basis vectors $b(\tilde{x})$ for the CATE estimation we construct cubic b-splines with $5$ degrees of freedom in the one dimensional case and a tensor product of quadratic b-splines with $5$ degrees of freedom in the two-dimensional case.
Appendix \ref{app:rob} contains evaluation and sensitivity checks with different learners and policies based on the confidence intervals.

Using the estimated score elements $\psi_b(W_i,\hat{\eta})$ from the \texttt{DoubleML}-package, we can apply weighted classifiers to obtain the policy learning results. 
As mentioned by \citet{athey2021}, greedy classification trees will not achieve the optimal regret bound as they do not optimize properly over the whole policy space. Consequently, we rely on the \texttt{R}-package \texttt{policytree} (\cite{Sverdrup2020}) to estimate the policy with exact tree search (results using greedy trees are available in Appendix \ref{app:addlearner}).

To illustrate the effect of different thresholds, we consider $\gamma\in\{0.01,0.03,0.05\}$ and report the corresponding policy. 

\begin{figure}[ht]
    \centering
    \includegraphics[width=\linewidth]{fig/all_policies_l3.png}
    \caption{Comparisons of the estimated policies.}
    \label{fig:policies}
\end{figure}

As it is shown in Figure \ref{fig:policies}, all methodologies lead to a major split along the main color measurement. This is in accordance to the domain knowledge described in Section \ref{Sec:intro}. While the one-dimensional CATE policy and the depth-1 tree are not flexible enough to display splits along the secondary color measurement, the two-dimensional CATE and the depth-2 tree split the area further. Nevertheless, all methodologies lead to similar decision rules. 

\begin{table}[h!]
\centering
\begin{tabular}{lrrr}
\toprule
Method       &  Policy 0 &  Policy 1 &  Policy 2 \\
\midrule
CATE 1D      &    0.3864 &    0.3352 &    0.2806 \\
CATE 2D      &    0.4179 &    0.3599 &    0.2973 \\
Depth-1 Tree &    0.4309 &    0.3842 &    0.2555 \\
Depth-2 Tree &    0.4708 &    0.4222 &    0.2393 \\
\bottomrule
\end{tabular}
\caption{Share of reworked panels per policy estimate (Share of treated in the observations: $0.2101$).}
\label{tab:policies}
\end{table}

Table \ref{tab:policies} contains the share of observations that would have to be treated under the estimated policies.
Even under the largest threshold ($5\%$) there would be a significant increase in workload from reworking, which nearly doubles for $\gamma=0.01$. This suggests that, neglecting capacity restrictions, there could be improvements in overall yield by employing a data-driven rework policy.

To evaluate the average effect of the different policies, we estimate the group average treatment effect (GATE) for the reworked panels under each policy (ATTE for the observational data or current policy).

\begin{table}[h!]
\centering
\begin{tabular}{lrrr}
\toprule
Method &  Policy 0 &  Policy 1 &  Policy 2 \\
\midrule
CATE 1D      &    0.0748 &    0.0790 &    0.0880 \\
CATE 2D      &    0.0717 &    0.0811 &    0.0867 \\
Depth-1 Tree &    0.0699 &    0.0755 &    0.0949 \\
Depth-2 Tree &    0.0686 &    0.0738 &    0.1026 \\
\bottomrule
\end{tabular}
\caption{GATE by suggested treatment groups of the policy estimation (GATE in observed policy: $0.0903$).}
\label{tab:gate}
\end{table}

Surprisingly, the current observational policy still has a very high average effect compared to the estimated policies. Based on our analysis, the average treatment effect on the group of to-be-treated chip lots improves only for the $5\%$-threshold tree policies. However, a smaller average effect per panel is not necessarily disadvantageous, as the share of treated is increasing. 

To quantify the potential improvement in overall yield with respect to the corresponding policy, we weight the GATE with the share of reworked panels.

\begin{table}[h!]
\centering
\begin{tabular}{lrrr}
\toprule
Method &      0 &      1 &      2 \\
\midrule
CATE 1D      & 0.0289 & 0.0265 & 0.0247 \\
CATE 2D      & 0.0300 & 0.0292 & 0.0258 \\
Depth-1 Tree & 0.0301 & 0.0290 & 0.0242 \\
Depth-2 Tree & 0.0323 & 0.0312 & 0.0246 \\
\bottomrule
\end{tabular}
\caption{Value of the estimated policies (Value of observed policy: $0.0178$).}
\label{tab:value}
\end{table}

The results in Table \ref{tab:value} suggest a large improvement in policy value for all data-driven policies. The improvement is mainly driven by a larger fraction of reworked panels. Even if capacity constraints do not allow for a rework of $40\%$, e.g. the one-dimensional CATE suggests, that $0.62\%$ more yield by reworking could be achieved by reworking $6.62\%$ additional chip lots.

%% file: sections/conclusion.tex
\section{Conclusion}
In this paper, we have used causal machine learning to estimate different rework policies for the color conversion process in opto-electronic semiconductor manufacturing.
The robustness and similarity of the different approaches supports the original argument, that the first principal component of the $C_1$ measurement contains most of the relevant information for a rework decision. This highlights the idea, that domain knowledge can be very helpful to design simple but effective data-driven policies.
Consequently, to improve the current rework decision rule, we propose a policy where the treatment decision is based on the main color point measure $C_m$. This would increase the number of reworked panels and still select the panels where a rework is the most effective, increasing the overall yield. The more flexible policy learners such as the two-dimensional CATE and the depth-2 tree also suggest splits along the secondary axis. However, we are cautious about these as they might as well be a product of overfitting. As visible in Figure \ref{fig:pca_obs}, the overlap of the data is mostly limited to the interval of $C_{m} \in [-0.005,0.005]$. Outside this area, predictions overemphasize single data points, leading to overfitting. The confidence intervals in these regions are considerably large (including zero), which underlines the substantial amount of statistical uncertainty within these regions.

In conclusion, it is very important to combine implications from causal inference with the knowledge of domain experts to make value out of the data. Carefully weighting up the possible cut-offs and also taking into account how many reworks are feasible with production capacities, we suggest implementing a clear cut-off along the main measurement which allows an easy implementation and thus higher compliance of the decision makers. Furthermore, at a cut-off, the real effect could be then measured with techniques such as regression discontinuity to allow for further analysis.

%% file: sections/appendix/robustness_L3.tex
\section{Robustness checks}
\label{app:rob}
\subsection{Additional Policy Learning Results}\label{app:addlearner}

In this section, we present additional results for policy learning.
The policy estimation with greedy decision trees relies on the \texttt{DecisionTreeClassifier} from \texttt{scikit-learn} (\cite{scikit-learn}). We also include a weighted classification with a rbf kernel in a support vector machine. However, we caution that the theoretical bonds by \citet{athey2021} do no apply to this learning method, since they are based on bounds on the VC-dimension  of the policy class $\Pi$.

\begin{figure}[ht]
    \centering
    \includegraphics[width=\linewidth]{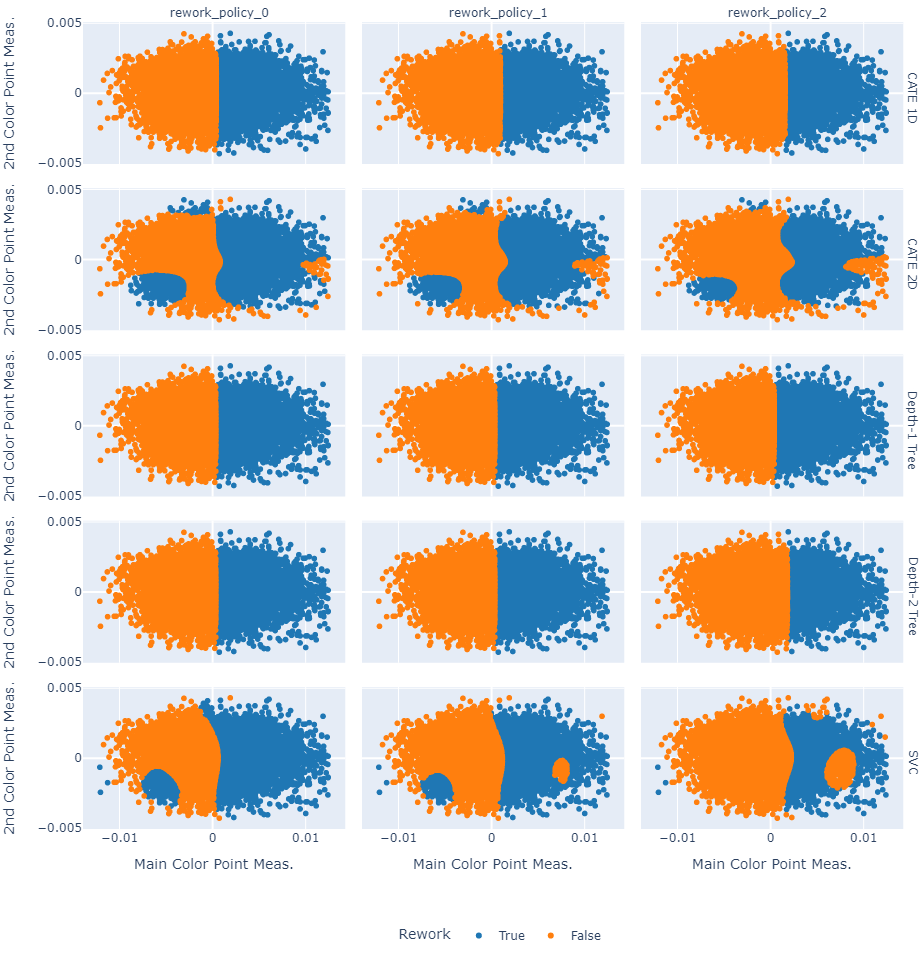}
    \caption{Comparisons of the estimated policies with additional methods.}
    \label{fig:pol_lgbm}
\end{figure}

Figure \ref{fig:pol_lgbm} contains the resulting policies for the tuned ML LGBM learner as in the main analysis. We see, that the greedy tree search algorithm finds no splits along the secondary color measurement. The rbf kernel encircles little areas of color point measurements that have a higher value. The areas are very similar to the two-dimensional CATE. However, they are in an area with few observations, so the evidence from them is limited.

\FloatBarrier
\subsection{DoubleML Machine Learner Comparisons}
In this section, we explore the robustness of our estimates by varying the ML method which is used to construct the estimators in Equations \ref{eq:g_0} and \ref{eq:m_0}. We compare the identical policy learning approaches as in Appendix \ref{app:addlearner}. The comparison includes results for XGBoost, random forest and linear learners. Table \ref{tab:rmse} contains the $RMSE$ of the respective ML methods in the nuisance estimation. 

\begin{table}[h]
\centering
\begin{tabular}{lrrr}
\toprule
Learner &  RMSE $m(x)$ &  RMSE $g_0(x)$ &  RMSE $g_1(x)$ \\
\midrule
Tuned LGBM    &    0.2821 &    0.8862 &    1.001 \\
XGBoost       &    0.2857 &    0.9078 &    1.043 \\
Random Forest &    0.2983 &    0.9581 &    1.077 \\
Linear Models &    0.2925 &    0.9682 &    1.083 \\
\bottomrule
\end{tabular}
\caption{RMSE for different nuisance learners.}
\label{tab:rmse}
\end{table}

\begin{figure}[ht]
    \centering
    \includegraphics[width=\linewidth]{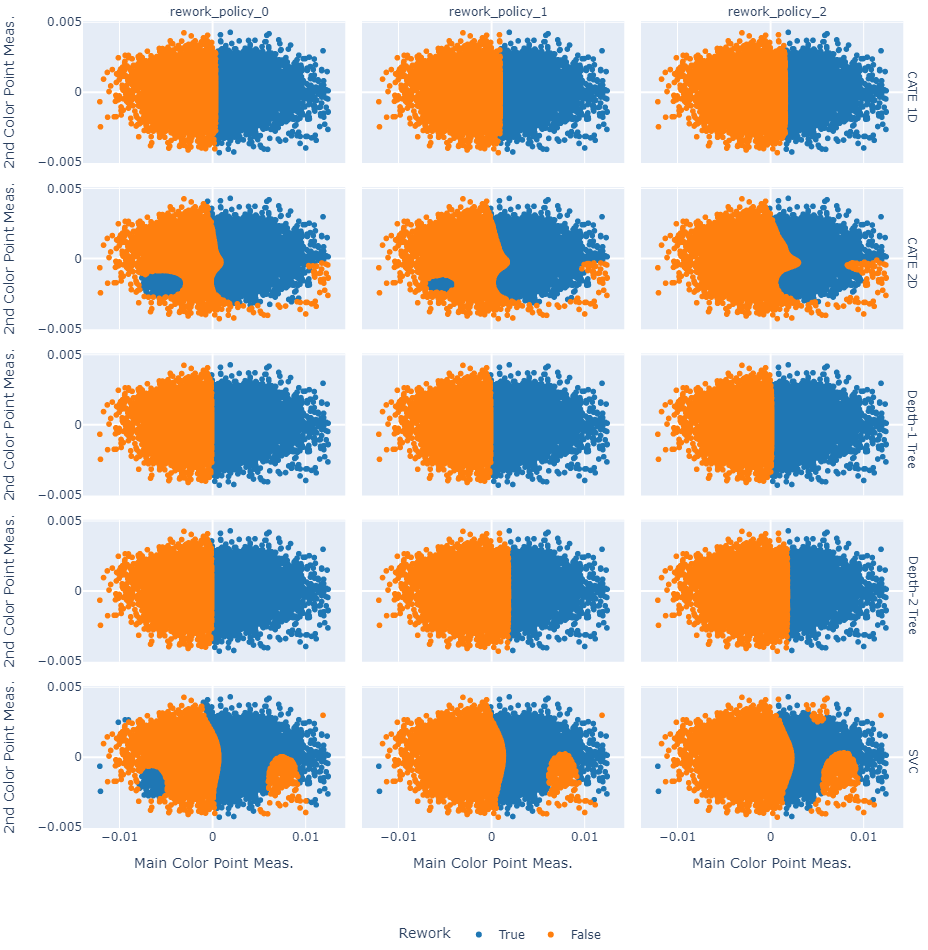}
    \caption{Comparisons of the estimated policies with the nuisance components estimated by gradient boosting implementation \texttt{xgboost}.}
    \label{fig:pol_xg}
\end{figure}

\begin{figure}[ht]
    \centering
    \includegraphics[width=\linewidth]{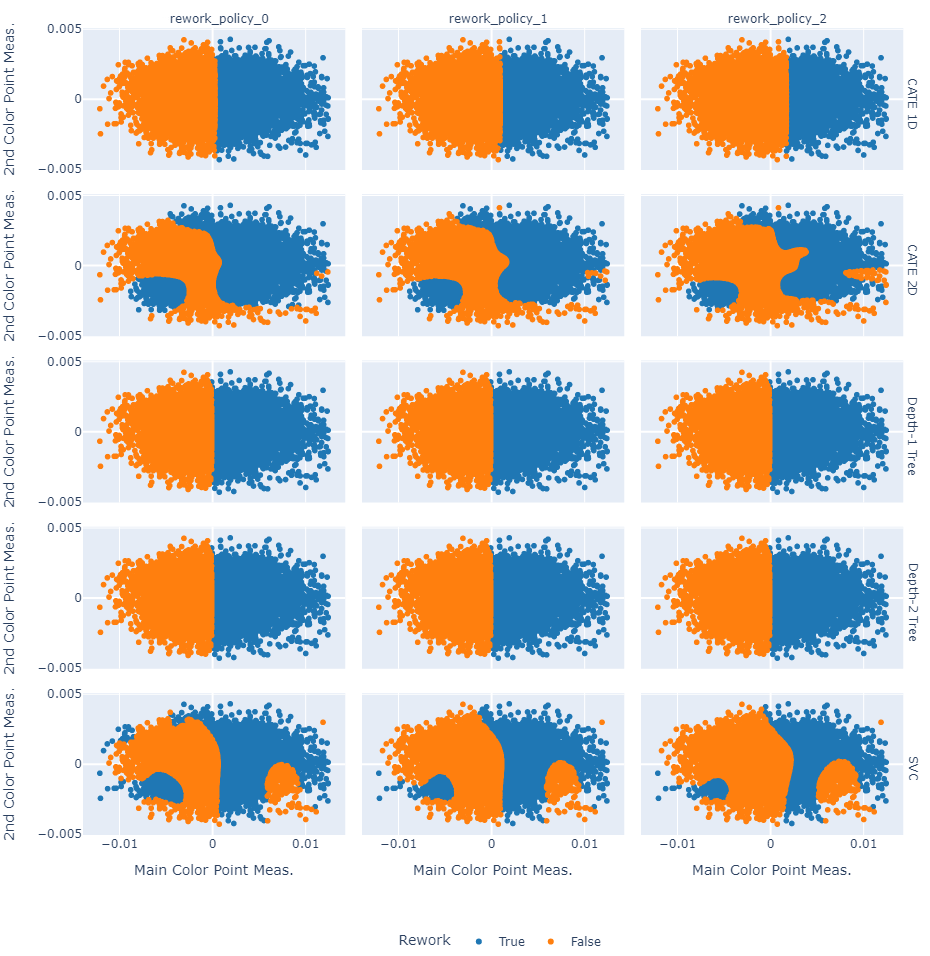}
    \caption{Comparisons of the estimated policies with the nuisance components estimated by the random forest implementation of \texttt{scikit-learn}.}
    \label{fig:pol_rf}
\end{figure}

\begin{figure}[ht]
    \centering
    \includegraphics[width=\linewidth]{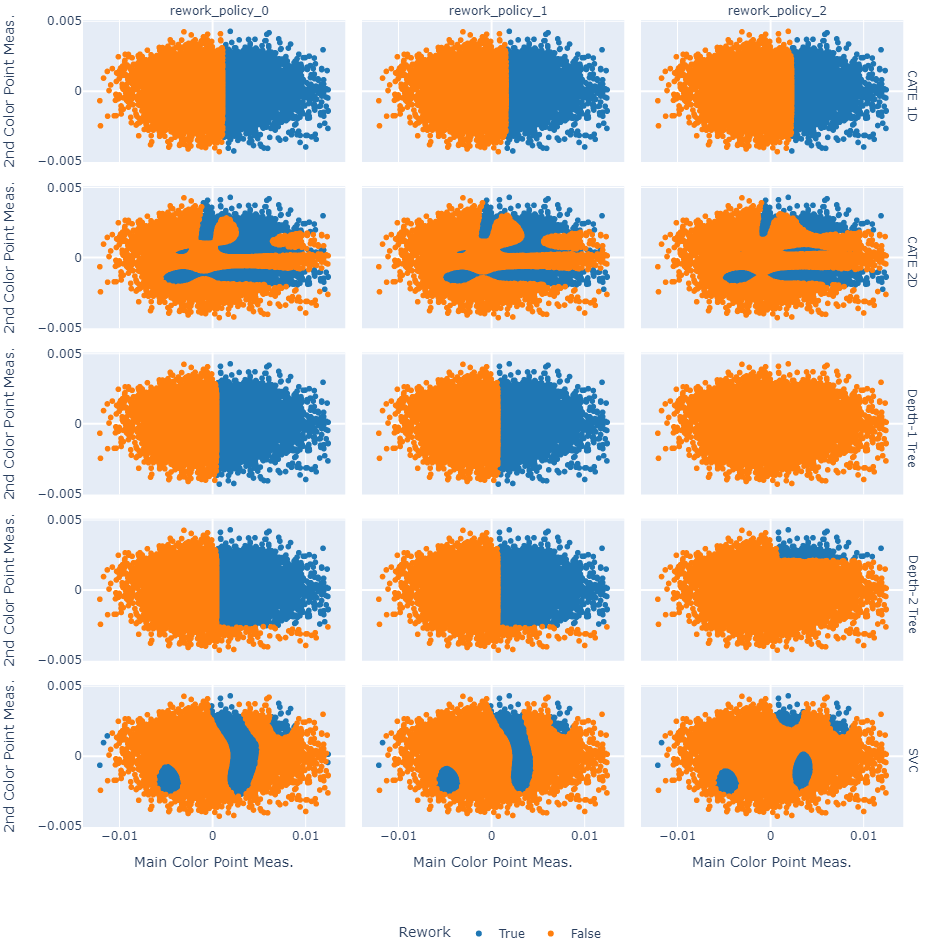}
    \caption{Comparisons of the estimated policies with the nuisance components estimated by linear models.}
    \label{fig:pol_lin}
\end{figure}

Both gradient boosting frameworks achieve good predictive performance. Random forest and linear model show a slightly weaker performance. In Figures \ref{fig:pol_xg},  \ref{fig:pol_rf} and \ref{fig:pol_lin} we show the results for the policies (to allow comparisons to the learner in the main analysis in Figure \ref{fig:pol_lgbm}). The results seem to be rather robust across multiple non-linear learners, especially in the main component split. For the linear models, we see a larger deviation in the policies, which, combined with the higher nuisance $RMSE$ might indicate that linear models are not sufficient to estimate all, potentially complex, dependencies in the data.

\FloatBarrier

\subsection{Policies Based on Confidence Intervals}\label{sec:confint}

As additional results, we present policies based on the conditional average treatment effect lower confidence interval. We construct a pointwise-valid confidence interval via a gaussian multiplier bootstrap as proposed in \citet{semenova2021debiased} at the level $2\alpha$. We then evaluate the lower bound to derive a rework cut-off, for which we are certain that the effect is above the desired threshold $\gamma \in \{0.01,0.03,0.05\}$ with probability $1-\alpha$. We choose $\alpha=5\%$.

Table \ref{tab:robshare} displays the share of treated under the conservative policies. We see, that compared to Table \ref{tab:policies}, less chip lots are assigned to be reworked. 

\begin{table}[h!]
\centering
\begin{tabular}{lrrr}
\toprule
Method &  Policy 0 &  Policy 1 &  Policy 2 \\
\midrule
CATE 1D &    0.3446 &    0.2754 &    0.2067 \\
CATE 2D &    0.2152 &    0.0852 &    0.0459 \\
\bottomrule
\end{tabular}
\caption{Share of treated chip lots under the conservative policies.}
\label{tab:robshare}
\end{table}

Unsurprisingly, this leads to an increase of GATE (Table \ref{tab:robgate}) in the group of treated, compared to Table \ref{tab:gate}.

\begin{table}[h!]
\centering
\begin{tabular}{lrrr}
\toprule
Method &  Policy 0 &  Policy 1 &  Policy 2 \\
\midrule
CATE 1D &    0.0758 &    0.0862 &    0.0967 \\
CATE 2D &    0.0797 &    0.0950 &    0.1328 \\
\bottomrule
\end{tabular}
\caption{Grouped average treatment effect for the group of treated chip lots under the conservative policies.}
\label{tab:robgate}
\end{table}

In the case of the one-dimensional CATE, we believe this is a possibility to make the cut-off even more conservative and still increase overall yield. In the case of the two-dimensional CATE, we are careful in interpretation, as using the confidence interval might overfit to a large extend on small areas with little observations but a potentially high value. The values of these policies are displayed in Table \ref{tab:robvalue} (in comparison to Table \ref{tab:value}).

\begin{table}[h!]
\centering
\begin{tabular}{lrrr}
\toprule
Method &      0 &      1 &      2 \\
\midrule
CATE 1D & 0.0261 & 0.0237 & 0.0200 \\
CATE 2D & 0.0171 & 0.0081 & 0.0061 \\
\bottomrule
\end{tabular}
\caption{Value of the conservative CATE policies.}
\label{tab:robvalue}
\end{table}

\begin{figure}[ht]
\includegraphics[width=\linewidth]{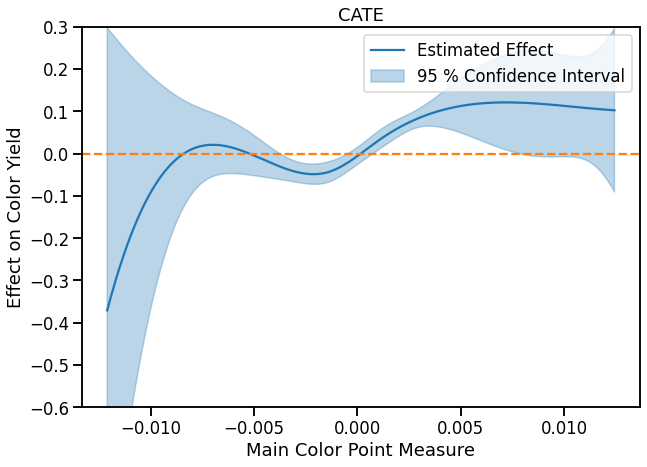}
\caption{CATE over the main color measurement for the full space of $C_m$ values.}\label{fig:full_cate}
\end{figure}

Additionally, we illustrate in Figure \ref{fig:full_cate} the influence of the little overlap outside the interval $C_m \in [-0.005,0.005]$. While the CATE is approximately constant, the intervals becoming increasingly large, making the analysis more difficult due to statistical uncertainty.

\FloatBarrier

%% file: sections/appendix/results_L4.tex
\section{Results for Product Type ``T2''}
\label{app:L4}
In this Section, we will present the analysis for product type ``T2''. Overall, the results are very similar, as the process is the same but only the product properties differ to some extend. Also in this setting, we would suggest a cut-off along the main component.
\subsection{Results}
\begin{figure}[ht]
    \centering
    \includegraphics[width=\linewidth]{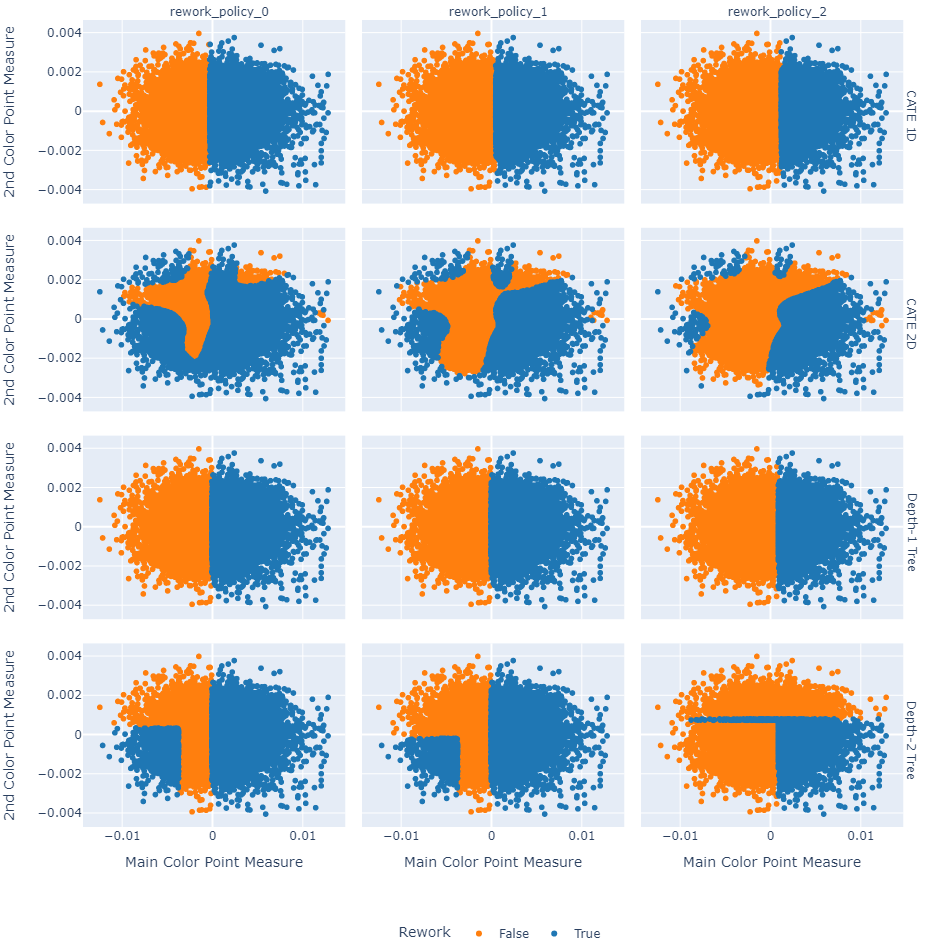}
    \caption{Comparisons of the estimated policies for product type ``T2''.}
    \label{fig:policies_l4}
\end{figure}

\begin{table}[H]
\centering
\begin{tabular}{lrrr}
\toprule
Method &  Policy 0 &  Policy 1 &  Policy 2 \\
\midrule
CATE 1D      &    0.4848 &    0.4087 &    0.3455 \\
CATE 2D      &    0.5106 &    0.3856 &    0.3127 \\
Depth-1 Tree &    0.4923 &    0.4907 &    0.3708 \\
Depth-2 Tree &    0.5624 &    0.4794 &    0.3013 \\
\bottomrule
\end{tabular}
\caption{Share of treated under the estimated policies for product type ``T2'' (Share in the observations: $0.2472$).}
\end{table}

\begin{table}[H]
\centering
\begin{tabular}{lrrr}
\toprule
Method &  Policy 0 &  Policy 1 &  Policy 2 \\
\midrule
CATE 1D      &    0.0818 &    0.0880 &    0.0961 \\
CATE 2D      &    0.0777 &    0.0950 &    0.1112 \\
Depth-1 Tree &    0.0819 &    0.0821 &    0.0951 \\
Depth-2 Tree &    0.0772 &    0.0865 &    0.1128 \\
\bottomrule
\end{tabular}
\caption{GATEs for the estimated policies for product type ``T2''. (GATE in the group of observed reworked: $0.1040$).}
\end{table}

\begin{table}[H]
\centering
\begin{tabular}{lrrr}
\toprule
Method &  Policy 0 &  Policy 1 &  Policy 2 \\
\midrule
CATE 1D      & 0.0397 & 0.0360 & 0.0332 \\
CATE 2D      & 0.0397 & 0.0366 & 0.0348 \\
Depth-1 Tree & 0.0403 & 0.0403 & 0.0353 \\
Depth-2 Tree & 0.0434 & 0.0415 & 0.0340 \\
\bottomrule
\end{tabular}
\caption{Value of the estimated policies for product type ``T2'' (Value in observed policy: $0.0036$).}
\end{table}